# From ADP to the Brain: Foundations, Roadmap, Challenges and Research Priorities

Dr. Paul J. Werbos
Engineering Directorate
National Science Foundation*
Arlington, Virginia, US
pwerbos@nsf.gov

*Abstract*—This paper defines and discusses "Mouse Level Computational Intelligence" (MLCI) as a grand challenge for the coming century. It provides a specific roadmap to reach that target, citing relevant work and review papers and discussing the relation to funding priorities in two NSF funding activities. – the ongoing Energy, Power and Adaptive Systems program (EPAS) and the recent initiative in Cognitive Optimization and Prediction (COPN). It elaborates on the first step, "vector intelligence," a challenge in the development of universal learning systems, which itself will require considerable new research to attain. This in turn is a crucial prerequisite to true functional understanding of how mammal brains achieve such general learning capabilities.

## I. Introduction

The Energy, Power and Adaptive Systems (EPAS) program at NSF welcomes proposals from a wide variety of topics discussed in this workshop, from neural networks to control theory to prediction and system identification and operations research, and applications to areas from energy to vehicles to biological modeling, robotics, and so on. However, given the acute limitations of funding for electrical engineering as a whole, priority is given to proposals with the greatest potential impact on the big picture. There are many areas in which basic R&D could have a big impact. This paper will focus on one of the most important areas.

In about half of the panels which I run I ask the panel: "Which of these proposals, if funded, would have the biggest impact on the probability that we get to the target of 'Mouse Level Computational Intelligence' (MLCI) at the soonest time?" This paper will specify that target more precisely, and discuss a roadmap for getting there, building on the great progress in adaptive approximate dynamic programming (ADP) and in "cognitive prediction" in recent years. It will give pointers to important sources of more detailed information.

It is crucial to remember that MLCI is a discrete target, a grand challenge, cutting across multiple disciplines.

## II. Qualitative Description of the MLCI Challenge

In broad terms, the grand challenge is to build a universal learning system which can learn to converge towards optimal policies "in any environment", at least as well as the brain of a mouse can. This is discussed in more detail in the NSF announcement for Cognitive Optimization and Prediction (COPN) [1], which was the outcome of extensive discussions across the Engineering Directorate of NSF. In essence, the idea is to understand and replicate this particular kind of general purpose learning ability, at the level of what the basic mammal brain can do.

NSF did not assert that optimization across time is the only principle active in the mammal brain; however, the brain of the mouse does seem to have some ability to learn how to maximize its probability of survival as it crosses fields of great unknown risk, as it also seeks more positive payoffs (like food). It does not adhere to robust control; there is no way to guarantee its survival in an absolute way, in the challenging real world of its environment. However, it does possess a high degree of "resilience" – ability to minimize the probability of disaster

## III. Initial Mathematical Formulation and the Link to Cognitive Prediction

The challenge here is to build a learning system which achieves a certain level of general performance across all ADP tasks. There are several ways to formulate exactly what an ADP task is; here I will pick a minimal, discrete time version of the ADP task which is nonetheless difficult and broad enough to capture the full challenge.

Let us begin by assuming that our ADP learning system is asked to maximize the expected value across time of a known utility function U($\mathbf{Y}$), based solely on observations of $\mathbf{Y}$(t) over time t and of its own actions $\mathbf{u}$(t), where $\mathbf{Y}$(t) is governed by an unknown stochastic process of the form:

$$Y(t) = h(X(t), E_1(t)) \quad (1)$$
$$X(t) = f(X(t-1), u(t-1), E_2(t)), \quad (2)$$

where **X**, **Y**, $E_1$, $E_2$ and **u** may be collections of discrete variables, or may be vectors in $R^n$ (i.e. collections of continuous variables), a hybrid combination of the two, or a collection of discrete and continuous variables defined over a graph or a physical space (as in PDE control). $E_1$ and $E_2$ are collections of "random variables," or, to use more universal terminology, "i.i.d. random variables."

Of course, the learning task is different, depending on which type of object we choose for X and Y. A major practical difficulty in ADP research is the need for better communication between the research groups which make one choice or another. The growth in ADP research in different communities has made it ever more challenging to maintain a unified understanding, which integrates what is being learned in these communities; see [2] for a current survey and effort at integration. The MLCI challenge asks us to come up with the most power possible general-purpose learning system, which tends to require that we bring together the fundamental principles, capabilities and insights important to all these more specific areas. It also requires the development of algorithms which fully exploit the computer power of new, massively parallel chips and systems [3,4], analogous to the massive parallelism of the brain.

Sections IV and V will define the challenge more precisely. For now – the challenge is to develop the most powerful possible learning system for the general case, where **X** and **Y** are sets of discrete and continuous variables which may be formally described as vectors, but which assume symmetry properties which effectively account for and include control over a graph and control over partial differential equations (PDE) as special cases. This is essentially the same as the "cognitive optimization" challenge put forth in [1].

To meet the MLCI challenge, we must also succeed in addressing a related challenge, which NSF has called "cognitive prediction" [1]. Crudely, cognitive prediction is the task of building the most powerful possible universal learning system to learn the functions f and h, or to predict $Y(\tau)$ for future times $\tau > t$ based on knowledge of $Y(t)$, $Y(t-1)$, etc., or to estimate $X(t)$ or $Pr(X(t))$ or a condensed representation of $Pr(X(t))$. In control theory, this is sometimes called "the system identification task." But here again, it is important to draw on what can be learned from other areas, such as statistics, neural networks, signal processing and machine learning.

Development and use of cognitive prediction is important to ADP in many ways. For example, in the general case, where X is not directly observed, systems which try to learn the optimal policy or value function as a function of Y can be grossly suboptimal; this leads to major performance problems in real world applications [1]. This can be overcome by using policy and value function approximators which include a condensed representation of $Pr(X(t))$ as part of their input [1]; thus we can use a "cognitive prediction" module within our ADP system, to give it more general capability.

As another example – the ability to approximate a nonlinear value function is crucial to the power of any general ADP system. "Value function approximation" is often seen as the big challenge to ADP in practical problems today. How is it that some people are unable to learn usable value function approximators as a function of 30 input variables, while the mouse brain can cope with thousands or even millions of variables? There has been enormous progress in the area of cognitive prediction in recent years [5], as people have begun to demonstrate an ability to handle thousands or millions of input variables, by using modified types of neural network design, which are also applicable in ADP. Of course, there have been a few special cases in ADP where piecewise linear value function approximators have been good enough for some large problems, but for the general case (the MLCI challenge) we need better.

Some further examples were also given in my keynote talk on this subject at SSCI-2013, which is planned to be posted at the website education.ieee-cis.org.

## IV. DEFINING THE CHALLENGE AND THE ROADMAP MORE PRECISELY

Certain large corporations, funded by DARPA, have announced that "We have already built the equivalent of a cat brain in computer hardware." But in fact, they only built hardware to simulate some current models of how neurons work, with enough neurons to match the cat, but without any ability to solve complex optimization problems or the complex problems of everyday life which cats and mice learn to solve.

Nature itself did not learn to build a mouse overnight. There is a fascinating progression in nature from less general and less powerful types of brains, on up to the general mammal brain [6]. Many of us are also interested in levels of intelligence beyond that of the mouse brain [7]; however, the grand challenge in science for this century (and perhaps the next) is to get a full understanding and ability to replicate the level of general intelligence we see in the mouse.

In [6], I propose a roadmap based in part on what we see in evolution and based in part on the key challenges in the ADP field. I described this as four major stages, like a ladder we must climb:

(1) "vector intelligence," in which we treat X and Y as vectors in $R^n$, **x** and **y**, and we do not assume special symmetry across components of those vectors;
(2) "spatial intelligence," in which we learn and fully exploit symmetry relations within **x** and **y**, accounting for example for the fact that they may contain arrays of pixels like images, so as to be able to handle a much larger number of component variables;

(3) "temporal intelligence," where we exploit modified Bellman's equations such as what I reported here in past years [8] to handle multiple time-intervals, in a far more powerful way than the simple ideas about "options" and skill learning now used in robotics;

(4) "creative" or "full mouse" intelligence, which addresses the problem of "brain-like stochastic search" by applying its spatial mapping capabilities to the task of mapping possible decisions.

Many already have cartoon designs today in artificial intelligence to try to capture these kinds of ideas, but we will be lucky if we learn to build full, optimal learning machines up to level (4) within a century from now. More precisely, it looks to me as if it would take another 20-25 years at the present rate, or more, to fully master optimal vector intelligence, let alone the higher levels which would build upon it. Certainly much more mathematical work needs to be done to master level 1, and there are many useful applications yet to be explored with the exciting new work which begins to capture the possibilities at level 2. I was hoping to fund more work aimed at level 3 a few years ago, but many of the practical applications required mastery of (1) and (2) first.

Perhaps in principle I should have defined a "level 1.5" which is like vector intelligence, but addresses hybrid systems, a mix of continuous and discrete variables. Likewise, there is a lot we can learn from "level 0" intelligence, in which **X** and **Y** are just collections of discrete variables, and f and h are finite-state automata.

There is very important research to be done now aimed at level 2, which is critical to applications like large-scale electric power systems and streaming video; however, for reasons of length, I ask those readers who are interested to go to the many sources cited in the various review papers cited above. The remainder of this paper will get deeper and more specific about the quest to achieve "vector intelligence," which has not yet been fully accomplished.

## V. CHALLENGES IN REACHING VECTOR INTELLIGENCE

What does it mean, precisely, to develop an "optimal universal learning system" for ADP systems or for prediction systems, at the level of vector intelligence?

Some theorists have argued that universal learning is impossible, based on arguments which cite the old adage "there is no free lunch." There are times when those arguments actually become quite humorous, if one studies them carefully [5]. Fortunately, the brain is living proof that some kind of universal learning ability is possible, and extremely important.

One way to define "universal learning" is to employ a key concept from statistics, with deep roots in philosophy, called "uninformative priors." One can define a specific universal learning task by augmenting section III with the assumption that the functions f and h are taken (sampled) at random from some prior distribution Pr(f,g). If that distribution is broad enough, and the class of possible functions f and h large enough, we may say that we have reasonably universal open-minded learning system.

This concept is already well-represented in the area of level-zero intelligent systems, in the study of prediction or optimization for finite state machines, where it is referred to as "Solomonoff priors" [5,9, 10]. If we taken $Pr(f,g) = e^{**(-kC(f,g))}$, where k is some constant and the complexity measure C equals the number of symbols required to express f and g as a program on a Turing machine, it can be proven that we get essentially the same results no matter what our choice of Turing machine; more precisely, if we "pick the wrong Turing machine," we lose the equivalent of no more than a finite number of extra symbols, so that there is only a finite, bounded amount of new information we need to learn to get back to the "right" Turing machine. These priors truly are universal, so long as we accept the basic principle of Occam's Razor, without which brain-like learning is clearly impossible.

Some in artificial intelligence would even say that this closes the issue. All we have to do is develop massive computer simulations of models sampled from this distribution, and test our competing learning designs against those simulations, in order to find out empirically through computation what is the best design. In fact, that general type of well-principled simulation study will be very important as part of developing the field of vector intelligence.

In practice, however, it is not so simple, for many reasons. We would like to have analytical results as well, and it is not so easy to derive analytical results for this class of priors. In addition, the Turing approach implicitly values symmetries (reuse of symbols), which is somewhat challenging, and really takes us up to a higher level. And finally, the world of continuous variables and stochastic disturbances warrants more explicit treatment, especially for those of us working with continuous variable problems.

One can approach "vector intelligence" as a ladder in itself, of ever more complex Pr(f,g), calling for more rigorous results, useful general software, and more systematic simulation work at each level of the ladder.

At a low stage of the ladder, we can consider the very simple special case where there is no vector **x**, no dependence on the past, and no partial observability, in prediction, such that we are simply trying to learn a function **y**=f(**u**,**e**). For that special case, Barron [11,12] has come close to offering an optimal universal learning rule, if we define the complexity measure C(f) as the Lipschitz measure of smoothness of f which Barron uses in his theorems. He has shown that the multilayer perceptron (MLP) can approximate smooth functions at the cost of many less parameters than the other commonly used methods, as the number of input parameters grows; the number of parameters is a key driver of the error in estimation and prediction. However, the use of suitable penalty functions to add to the error-function, and a limited use of "syncretism" [5], may make it possible to do better, even when we try to learn from a fixed database without the constraints of real-time learning and forgetting factors. Is it possible to build a supervised learning system for this low step of the ladder, which is provably as universal for this challenge as the Solomonoff priors theoretically are for theirs? That has yet to be proven. More work is needed to nail this down.

"Syncretism" is essentially a matter of using neighbor-based prediction (such as a low-cost approximation to local kernels) to predict the errors of a global forecasting model or function. It fits very well with Freud's picture of the "psychodynamics of ego and id.' [5]. Much work is needed to establish an optimal o(N)-cost family of universal learning designs in that space, taking into account the importance of memory constraints even in the brain itself.

Another low level of the ladder comes when we restrict f and g to the kind of functions which give us univariate stochastic processes, as in the classic text by Box and Jenkins [13]. Box and Jenkins essentially provide a universal (offline) learning system for that special case. Because it is universal, and widely disseminated, it is still the best standard tool used in prediction in many applications today. It does not use an explicit state space representation, as in equations 1 and 2, and thereby avoids the issues of nonuniqueness which plague much of the state-space work in the same case. However, one step up the ladder is the case of multivariate linear ARMA or ARMAX processes, for which a computationally efficient algorithm was not available until 1973/1974 [14]. The functions f and g representing univariate Box-Jenkins systems are a strict subset of the set of f and g representing multivariate ARMAX systems.
This is beautiful example of how universal learning works [15]; by choosing the more general class of models, we may pay a finite penalty when we are studying a system which really is univariate, but that penalty becomes ever smaller with experience [14], whereas if we choose a univariate model when studying a multivariate process, we will be forever far behind.

In ADP and adaptive control, there are exciting opportunities right now formore universal stability and performance results for stochastic and deterministic versions of multivariate ARMAX processes. Robust stability results related to these issues, not fully addressed in the current literature, were given in [16], and further discussed in my recent video course on ADP for the CLION center at the FedEx Institute of Technology.

Of course, the class of smooth functions f is a superset of the linear functions. Thus we may represent the general multivariate NARMAX model as:

$$\underline{y}(t+1)=f(\underline{y}(t),\underline{u}(t),\underline{R}(t),\underline{e}(t) \quad (3)$$
$$\underline{R}(t+1)=g(\underline{y}(t),\underline{u}(t),\underline{R}(t)) \quad (4)$$

and consider the challenges of prediction and optimization when f and g are taken from the set of smooth functions, exactly as in our discussion of Barron's work. I termed this class of system "time-lagged recurrent network (TLRN)" in publications decades ago [5], far antedating the narrow special cases discussed more often in today's machine learning. Since the multivariate ARMAX systems are a strict subset of this more general class, the TLRN provides more universal learning than multivariate ARMAX. Years ago, I consulted briefly with private companies who had been stuck in a kind of tug of war between simple MLP and univariate Box-Jenkins in their forecasting operations; when they shifted to TLRN, which is a superset of them both, and accounts for both kinds of complexity (nonlinearity and complex lag effects), they easily dominated both. TLRN are widely used in some large-scale crucial industrial activities, such as automotive applications and power generator control [17]; however, all the new work which is still needed for Barron's case is needed here as well. Note that adding additional time lags does not really add generality here, in this nonlinear multivariate situation

Standard TLRNs based on minimizing prediction error from time t to time t+1 is used in the high-powered general systems developed by Ford [18] and by Siemens [19]. These are perhaps the most powerful universal systems to learn to make predictions (in vector prediction) in the world today. They have done very well in forecasting competitions, and also have applications in control. However, extensive practical work on oprediction [14,20] has shown that there are some practical tricks which can lead to better results in limited circumstances; it will be important to extract the underlying theoretical principles, to apply them in a more general way, as in the examples of improved forecasting in chapter 10 of [21].

Finally, even within the world of vector prediction and control, it is important to consider the more general case where f and g themselves are not smooth functions, but functions defined by "an instantaneous recurrence" wrapped around a smooth function. In other words, we may still assume a kind of "smoothness-based complexity measure" for the probability distribution, not of f or g, but for the inner functions used when f and g are defined as Simultaneous Recurrent Networks (SRN). See [16, chapter 10] for a discussion of predictive networks which combine time-lagged recurrence and simultaneous recurrence in that way. Venayagamoorthy has done simulations illustrating how SRNs are a useful superset of MLPs, offering more universal learning ability [22].

Strictly speaking, there is yet a higher level of vector intelligence possible, if we consider a more explicit way of accounting for correlated uncertainties. In [16, chapter 13], I describe a Stochastic Encoder/Decoder/Predictor architecture, into which one can insert simpler types of recurrent networks as components. This design has been used on occasion in industry, as a kind of nonlinear version of factor analysis, but it becomes more important either with long time intervals or when looking for symmetries, as a step up towards spatial complexity [6].

## V. SUMMARY

Much more work will be needed before we can build, prove and make full practical use even of "vector intelligence," which is just one of the four big steps towards the MLCI challenge. Nevertheless, the "ladder" of challenges in vector control is well-defined, and we have the necessary starting points.

Just as theorems about universal learning with "Solomonoff priors" and finite state machines were proven decades ago, theorems and more general tools for universal vector intelligence should be possible in the coming 20-25 years. All the work which gets us there should be a high priority in research funding, and in new efforts to build bridges between engineering and cognitive neuroscience [1].